\documentclass{article}

\usepackage[english]{babel}

\usepackage[letterpaper,top=2cm,bottom=2cm,left=3cm,right=3cm,marginparwidth=1.75cm]{geometry}

\usepackage{amsmath}
\usepackage{graphicx}

\title{Rapid-Motion-Track: Markerless Tracking of Fast Human Motion with Deeper Learning}
\author{Renjie Li, Chun Yu Lao, Rebecca St. George,\\ Katherine Lawler, Saurabh Garg, Son N. Tran, Quan Bai, Jane Alty}

\begin{document}
\maketitle

% Objective, Materials and Methods, Results, Discussion, and Conclusion.
\begin{abstract}
\noindent \textbf{Objective}: 
The coordination of human movement directly reflects function of the central nervous system. Small deficits in movement are often the first sign of an underlying neurological problem. The objective of this research is to develop a new end-to-end, deep learning-based system, Rapid-Motion-Track (RMT) that can track the fastest human movement accurately when webcams or laptop cameras are used. 

\noindent \textbf{Materials and Methods}: 
We applied RMT to finger tapping, a well-validated test of motor control that is one of the most challenging human motions to track with computer vision due to the small keypoints of digits and the high velocities that are generated. We recorded 160 finger tapping assessments simultaneously with a standard 2D laptop camera (30 frames/sec) and a high-speed wearable sensor-based 3D motion tracking system (250 frames/sec). RMT and a range of DLC models were applied to the video data with tapping frequencies up to 8Hz to extract movement features.

\noindent \textbf{Results}:
The movement features (e.g. speed, rhythm, variance) identified with the new RMT system exhibited very high concurrent validity with the gold-standard measurements (97.3\% of RMT measures were within +/-0.5Hz of the Optotrak measures), and outperformed DLC and other advanced computer vision tools (around 88.2\% of DLC measures were within +/-0.5Hz of the Optotrak measures). RMT also accurately tracked a range of other rapid human movements such as foot tapping, head turning and sit-to -stand movements.

\noindent \textbf{Conclusion}: With the ubiquity of video technology in smart devices, the RMT method holds potential to transform access and accuracy of human movement assessment.
\end{abstract}

\section{Background and Significance}\label{sec1}
Assessment of human movement is critical to neuroscience as well as a broad range of other fields including sport science and rehabilitation medicine as it allows interrogation of brain networks, and measurement of the response to interventions~\cite{kim2011quantification,schneider2013influence,li2022applications}. Currently, movement features are evaluated by a clinician or researcher, using validated rating scales that provide ordinal scores for individual components such as speed and amplitude, or composite scores of these combined together~\cite{goetz2008updrs,heldman2011modified,mestre2018rating}. These methods are subjective, expensive (as it requires at least one rater, often a medical specialist, per participant) and imprecise, with considerable inter-rater variability~\cite{shirani2017finger}. To reduce bias, multiple raters are recommended to rate one participant's movement~\cite{shirani2017finger} but this is labor dependent, time consuming, and lacks the granularity of a continuous measure. Wearable sensors provide accuracy, but they require specialist equipment and tend to be expensive. Furthermore they cannot be applied at a population level for epidemiological or health screening purposes, and are largely inaccessible for people in remote areas. No wearable sensor methods have yet been routinely incorporated into standard clinical practice.

Due to the ubiquity of cameras in personal smart devices worldwide, deep learning-based computer vision methods applied to video recordings of human movement is a promising solution to overcome these limitations. However, current computer vision methods require cameras with a high sampling frequency that produce high resolution videos, and this limits the use of computer vision techniques with standard cameras in laptops or desktop webcams that have relatively low sampling rates. This is problematic as laptops and desktops are the most common method used in telemedicine~\cite{li2022moving} which has boomed in use since the COVID-19 pandemic.

The Finger Tapping Test (FTT) is a well validated test of human movement function. It is used clinically and in neuroscience research studies as a measure of fine motor control. Participants are required to tap their index finger against their thumb repeatedly and usually instructed to tap `as big and fast as possible'. The FTT is used to assess motor function across neuroscience research and an array of neurological disorders including the two most common neurodegenerative disorders: Parkinson’s disease~\cite{arias2012validity} and Alzheimer's disease~\cite{suzumura2016assessment}. With ageing populations, these two conditions already affect 61 million people worldwide and are predicted to affect 165 million by 2050~\cite{dorseyemerging}, so there is growing need for objective methods to evaluate the FTT. 

Our recent work, was the first to assess the performance of currently available cutting-edge computer vision models for extracting FTT movement features from relatively low frame per second (fps) web-cameras. This demonstrated that DeepLabCut (DLC) and other similar computer vision models (8 were assessed in total)~\cite{li2022moving} were able to reliably track finger tapping movements up to 4Hz. However, when FTT frequencies were above 4Hz, the motion blur on videos prevents accurate keypoint detection and renders DLC and other computer vision methods inaccurate. This means that whilst DLC techniques can track slower tapping speeds of patients with obvious motor impairment (e.g. in more advanced Parkinson’s), it would not be able to measure subtle changes in fast movements that are generated by young adult participants or by those in the pre-motor, or early, stages of Parkinson's and other neurological disorders, where movements are generally above 4Hz. This severely limits the application of computer vision methods in real world settings. Thus, there remains an urgent unmet need for a computer vision method that can  accurately measure human hand movements, and other fast human motions, using  standard webcam technologies. This would transform our ability to remotely assess humans in their own homes for both clinical and research purposes. With the global COVID-19 pandemic, this need is even more pressing. 

To overcome these shortcomings of DLC and other advanced computer vision techniques, we have developed a new system, called Rapid-Motion-Track (RMT) that can extract accurate features of fast human movements from standard (relatively low) 30fps laptop cameras. We designed and conducted a number of experiments whereby participants completed the FTT at a range of metronome- and self-paced frequencies. Tapping movements were simultaneously recorded with a standard laptop camera and high speed 3D wearable sensors (Optotrak, 250 fps). Features determined through RMT applied to the video recordings were validated against the gold standard, wearable sensor method and compared to DLC and other deep learning methods. 

Our main contribution is that the new RMT system can extract valid and accurate fast human movement features using relatively low fps cameras that are superior to current DLC and other computer vision models. Sub-contributions are:
\begin{itemize}
\item[$\bullet$]
We show RMT outperforms DLC and other state-of-the-art computer vision methods in two challenging cases: 1) tapping frequencies from 0.5Hz to 6Hz and 2) videos in low resolution ($256 \times 256$) and low frame rates (30 fps).
\item[$\bullet$]
We demonstrate that RMT has robust tracking across a range of human movements used in motor control assessments including sit-to-stand, head turning, foot tapping and leg agility tests.
\end{itemize}

\section{Materials and Methods}
\subsection{Data Collection}
The data collection process has been described in detail previously~\cite{li2022moving}. In brief, participants sat facing a laptop camera with an Optotrak high speed 3D camera behind them. The Optotrak system (Northern Digital Inc.) sensors (infrared Light Emitting Diodes) were attached on participant's index fingertip and thumb-tip (both on the back of the hand). Sixteen participants performed FTT for 20 seconds under 5 conditions (0.5 Hz, 1Hz, 2Hz, 3Hz and maximal speed (which was typically in the range 5-8 Hz)). The laptop camera recorded 2D videos and the Optotrak system recorded the real life positions of the two sensors. Further details can be found in our previous work~\cite{li2022moving}.

Each video contains around 600 frames ($20\times30$) and 20 frames were extracted from each video based on K-means clustering algorithm (K=10) for manually labelling the index fingertip and thumb-tip position. Total 4,400 (with finger digit position being labelled) frames were used for evaluating the digit tracking result. The hand movement features calculated based on Optotrak sensors were used as the feature ground truths for evaluating the feature extraction result.

\subsection{RapidMotionTrack System}
\begin{figure*}[ht]
%\vskip -.2cm
\centering
\includegraphics[width=\textwidth]{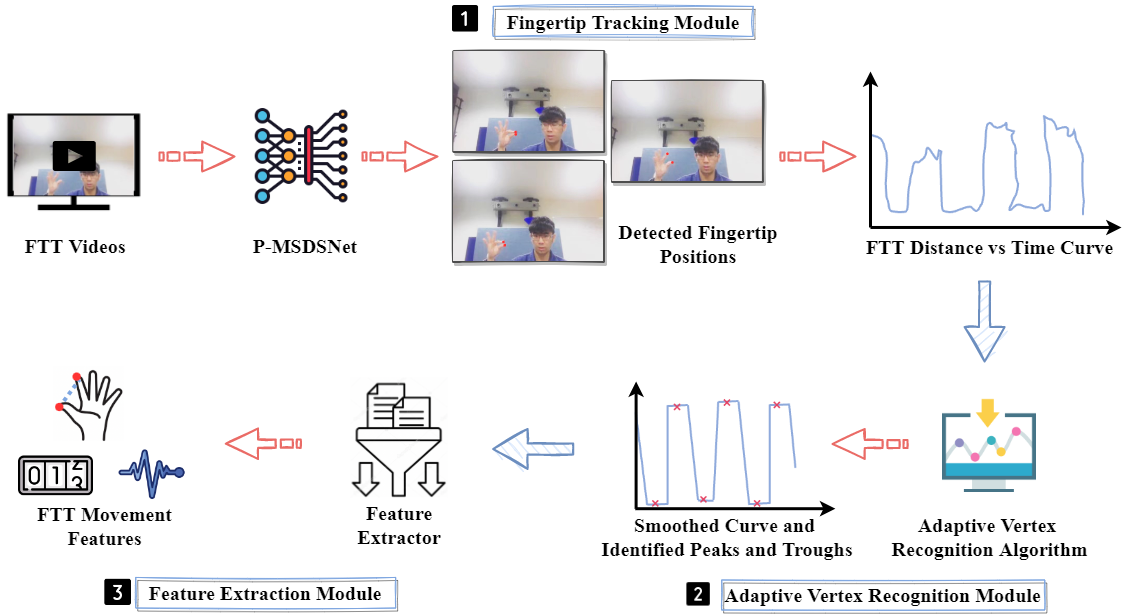}\\

\caption{An illustration of RapidMotionTrack system with three modules, including Fingertip Tracking Module, Adaptive Vertex Recognition Module and Feature Extraction Module.}
\label{fig:whole}
\end{figure*}

RapidMotionTrack takes 2D FTT videos as the input and the outputs are hand kinematic features. This system consists of three modules, i.e. Fingertip Tracking, Adaptive Vertex Recognition and Feature Extraction, shown in Figure~\ref{fig:whole}. The Fingertip Tracking module adapts our previous work P-MSDSNet~\cite{li2021parallel} to track thumb-tip and index fingertip positions on the 2D finger tapping videos. The role of the Adaptive Vertex Recognition module is to smooth the distance-versus-time curve adaptively based on the dominant tapping frequency of a participant to accurately localize the time point of tapping peaks and tapping valleys. The Feature Extraction module takes charge of extracting different hand movement features that would be useful for neuroscience research.

\subsubsection{Fingertip Tracking Module}
The Fingertip Tracking module takes 2D finger tapping videos as input and outputs a distance-versus-time curve graph. The distance is measured between the thumb-tip and the index fingertip. We applied and adjusted our previous work P-MSDSNet~\cite{li2021parallel} to track thumb-tip and index fingertip positions on each frame of the video. P-MSDSNet is a multi-scale neural network which can learn abundant features from high resolution feature maps to low resolution feature maps in a cyclical and cascade pattern. Due to P-MSDSNet’s deep supervision-based spatial attention mechanism on different scale levels of the input images, the network is able to extract discriminable features to effectively predict the locations of the thumb tip and the index fingertip.

\begin{figure*}[ht]
\centering
\includegraphics[width=\textwidth]{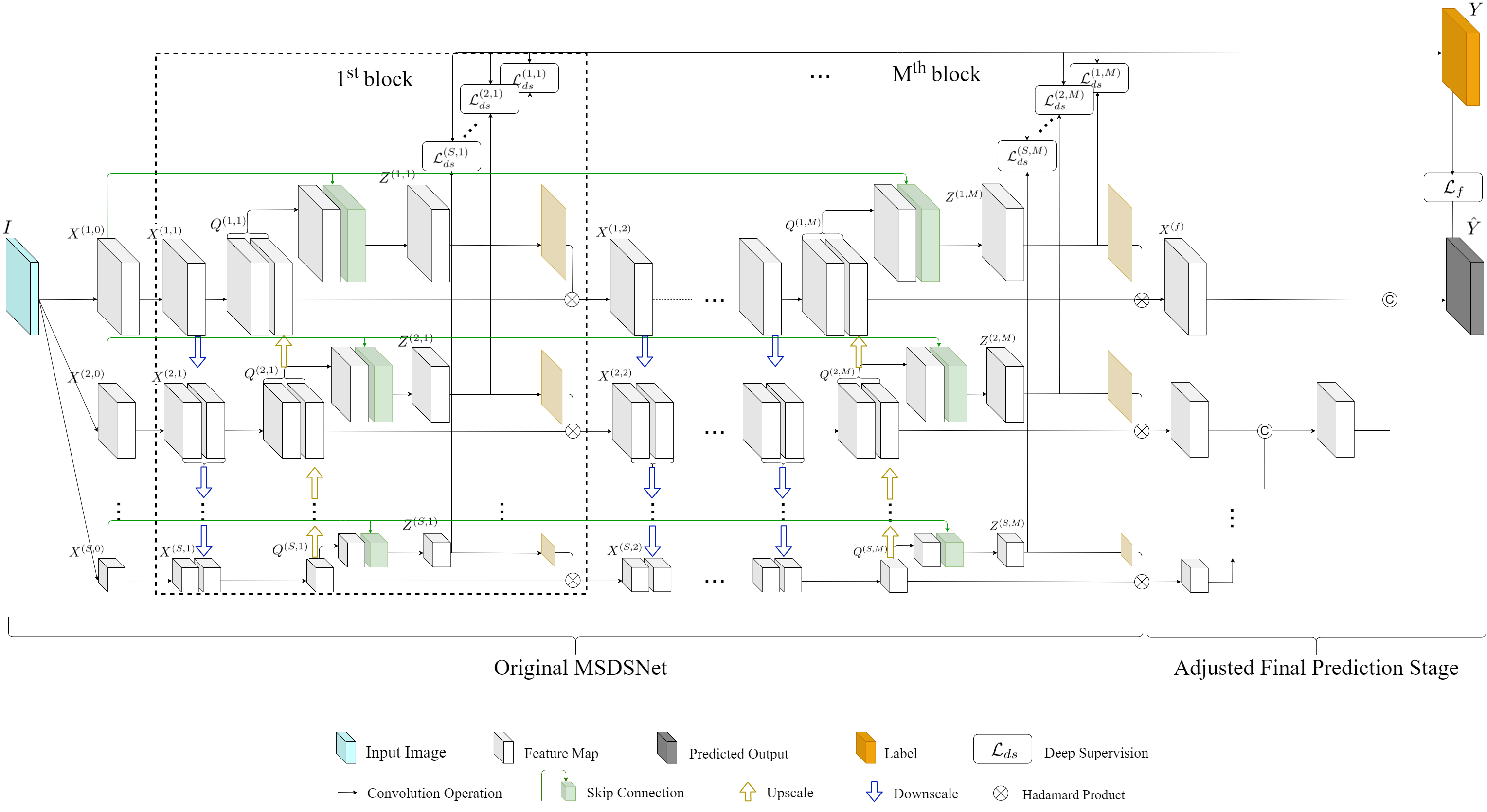}
\caption{An illustration of adjusted P-MSDSNet used in RapidMotionTrack. RapidMotionTrack applies 3 up-downscale blocks ($M=3$) and 5 different scale sizes ($S=5$). It also adds a final prediction stage to fuse the outputs from different scale feature maps together to make the final prediction.}
\label{fig:adjusted_msdsnet}
\end{figure*}

In the Fingertip Tracking module, we employed a P-MSDSNet with a stack of neural network blocks, each consisting of 5 different scaling operators. The multi-scale features are fused together in an up-down manner and are propagated to deeper levels under a deep supervision mechanism to increase the prediction accuracy. For FTT, some frames may have motion blur areas due to the very quick finger tapping. To mitigate the impact of the motion blur on the fingertips detection, we improve the final prediction stage network by taking into account the learned feature maps from different scales at the last neural network block in the stack (Figure~\ref{fig:adjusted_msdsnet}).

In the network forward propagation process, the original frame is transformed to 5 ($S=5$) different scale feature maps (from $X^{(1,0)}$ to $X^{(5,0)}$) through some convolutional operators. Then, along each scale, feature maps propagate information forward in a parallel manner, meanwhile, fuse different information through down-up scale connection blocks between adjacent scales. For example, $X^{(1,1)}$ feature map not only propagates forward along its own scale path, but also does the downscale operation and fuse with $X^{(2,1)}$ feature map. Then, $X^{(2,1)}$ feature map propagates forward after it fuses the downscaled information from $X^{(1,1)}$. In this case, information is spread not only along its self scale path but also across different scales paths in a downscale-upscale fusing mechanism. We apply 3 downscale-upscale blocks ($M=3$) and add deep supervision module at the end of each block~\cite{li2021parallel}. In the final prediction stage, we fuse the final feature map from the smallest scale to largest scale step by step using convolutional operator and concatenating operator. At last, the final prediction of heatmap is inferred at the largest scale $\hat{Y}$(the same scale with the original frame).

In the network training process, twenty frames were selected from each of the 220 videos by K-means ($K=10$) clustering algorithm~\cite{mathis2018deeplabcut} for manually annotating the positions of thumb-tip and index fingertip. Then, 4,400 annotated frames ($20 frames \times 220 videos$) were randomly split into training dataset (95\%) and testing dataset (5\%). We trained different deep learning models on training dataset and compared the performance on testing dataset. The annotated fingertip positions on each frame is transposed to a $H\times W\times 2$ heatmap, where $H$ and $W$ refer to the height and width of the frame and $2$ refers to the number of fingertips (in this case, thumb-tip and index fingertip). The heatmap is generated by a Gaussian function with $\boldsymbol{\mu} = \begin{bmatrix}
x  \\
y
\end{bmatrix}$ and $\boldsymbol{\Sigma} = \begin{bmatrix}
3 & 3\\
3 & 3
\end{bmatrix}$ for fingertip, $(x,y)$ is the position of the fingertip. For training, Mean Square Error (MSE) between the true fingertip position heatmap and predicted heatmap is used as loss function and Adam algorithm~\cite{kingma2014adam} is used as learning optimizer. The trained model will be used on each frame of each video to track the positions of thumb-tip and index fingertip.

After obtaining the positions of thumb-tip and index fingertip on each frame, we calculate the Euclidean distance between thumb-tip and index fingertip (in pixels) on each frame and draw the distance-versus-time curve graph.

\subsubsection{Adaptive Vertex Recognition Module}
Since the predicted fingertip position covers only one pixel area of the frame, even when the performance of fingertip tracking is good, it is inevitable that the distance vs time curve will not be smooth (this can be caused by little tremor when participant does the finger tapping). Unsmooth curve increases the difficulty in identifying peaks and troughs, which further leads to inaccurate feature extraction. To solve this issue, we developed an Adaptive Vertex Recognition algorithm to identify peaks and troughs along the distance-versus-time curve graph adaptively by filtering out rough areas.

\begin{figure}[ht]
\centering
\includegraphics[width=\textwidth]{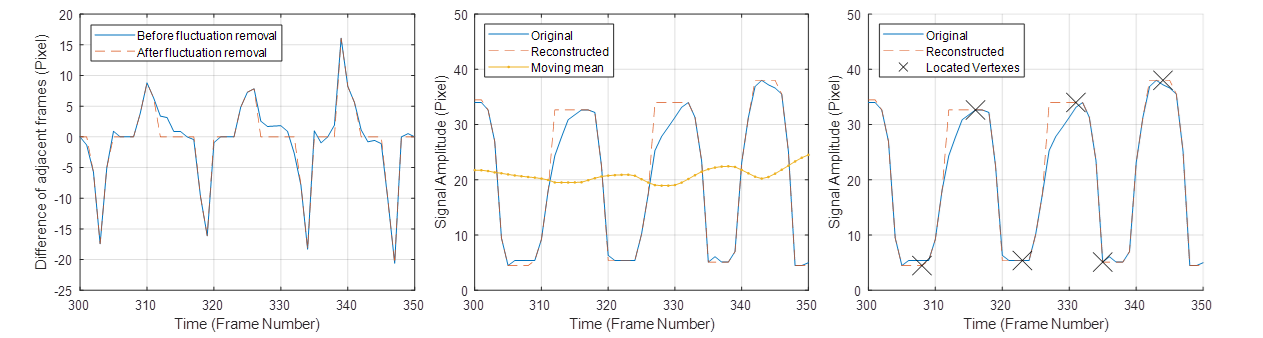}
\caption{Adaptive vertex recognition process.}
\label{fig:filter}
\end{figure}

Firstly, we use the distance difference of the adjacent frames ${\Delta S}$ as a criteria to remove the fluctuation of the signal. 
\begin{equation}
    \Delta S'=\left\{
    \begin{aligned}
    0 & & when & & \Delta S < \gamma_{flatness} R \\
    \Delta S & & when & & \Delta S \geq \gamma_{flatness} R
    \end{aligned}
    \right.
\end{equation}
where $S$ is the distance signal after time-averaged mean removal, $S'$ is the curve after fluctuation removal, $\Delta$ represents the difference between adjacent frames of specified signal, $\gamma_{flatness}$ is the threshold which set as 0.1 in the present study, and $R$ is the range of the interested signal. The distance signal before ($Delta S$) and after fluctuation removal ($\Delta S'$) are plotted side-by-side in Figure~\ref{fig:filter} (left).

Secondly, we reconstructed the signal based on $\Delta S'$ and $S$. The former obtains sharp change between sections (where $Delta S'=0$, i.e., finger open or close), the latter guides the magnitude of the reconstructed signal. Moreover, the existence of peaks and troughs can be determined by calculating the moving average value $\mu$ over the signal. Peaks are determined when the local section are above $\mu_{i}$, in contrast, troughs are recognised when the local sections are below.
\begin{equation}
\begin{aligned}
    &\mu_{i}= \tfrac{1}{n} \sum_{i-n/2}^{i+n/2}{S'_{i}} \\
    & \text{w.r.t } n= \gamma_{window} \times N
\end{aligned}
\end{equation}
where $\gamma_{window}$ is the relative window size (set as 0.1 in the present study) and $N$ is the total number of frames in the signal. Please note that the duration of the first and last platforms are omitted, since it often is meaningless waiting period. Figure~\ref{fig:filter} (center) shows the original signal $S$, reconstructed signal $S'$, and the moving mean $\mu$.

Due to the fact that the features, such as the time between each tap and open duration of the fingers, of every signal varies significantly. Thus, the cases of short and long opening are both need to be considered and processed separately to recognise the vertexes in a reliable manner throughout all videos, following are the formulas applied.
\begin{equation}
    t_{vertex}=\left\{
    \begin{aligned}
    t_{localvertex} & & when & & l_{section} \leq \gamma_{platform} N \\
    t_{median} & & when & & l_{section} > \gamma_{platform} N
    \end{aligned}
    \right.
\end{equation}
where $t_{localvertex}$ is the moment of the local maxima or minima of the section, $t_{median}$ is the central time of the local section, $l_{section}$ is the duration of the local section, and $\gamma_{platform}$ is the constant which set as 0.01 in the present study.

Once the time of local vertexes $t_{vertex}$ are located, the local height $S_{vertex}$ can be found by below:
\begin{equation}
    S_{vertex}=\left\{
    \begin{aligned}
    S_{localsection} & & when & & \Delta S'(t_{vertex}) = 0 \\
    S(t_{vertex}) & & when & & \Delta S'(t_{vertex}) > 0
    \end{aligned}
    \right.
\end{equation}
where $S_{localsection}$ is the maxima or minima of the local section. The sample signal, processed signal, and recognised vertexes are demonstrated in the Figure~\ref{fig:filter} (right). The curve of recognised vertexes ($t_{vertex}$, $S_{vertex}$) are ready for the next step, feature extraction in the following section.

\subsubsection{Feature Extraction}
The Feature Extraction module takes reconstructed distance (between thumb-tip and index fingertip)-versus-time graph data as the input and different hand kinematic features are the outputs. Based on key neuroscience measures, we summarized speed, amplitude and rhythm related features of the FTT and their calculations. These features are known to deteriorate in Parkinson's and other neurological disorders.

Speed related features include Mean Tapping Frequency (M-TF), Total Tapping Count (TTC), Maximum Speed (MS), Mean Inter Tap Interval (M-ITI) and Decrements on Speed (DoS). Amplitude related features include Coefficient of Variance of Amplitude (COV-A) and Decrements on Amplitude (DoA). Rhythm related features include Coefficient of Variance of Tapping Frequency (COV-TF) and Intra Individual Variability (IIV). Detailed validation results on all 9 features are presented in supplementary document.

%M-TF (Equation~\ref{equ:m-tf}) is the most widely used feature in analyzing FTT~\cite{jobbagy2005analysis}. It measures the tapping speed of a participant and has been validated as an important feature that has correlation with neurological disorders like Parkinson's disease. TTC (Equation~\ref{equ:ttc}) measures the number of tapping in a test, which is also an important feature in quantitative assessment of FTT in mild cognitive impairment, Alzheimer’s disease, and Parkinson’s disease~\cite{roalf2018quantitative}. M-ITI (Equation~\ref{equ:m-iti}) has been studied in~\cite{roalf2018quantitative} showing that AD and MCI groups produced fewer taps with higher M-ITI compared to health control. DoS (Equation~\ref{equ:dos}) measure the decrements of a participant's finger tapping in terms of speed.

%COV-A (Equation~\ref{equ:cov-a}) measures the variability of participants' tapping amplitude. DoA (Equation~\ref{equ:doa}) measures the decrements of a participant's finger tapping in terms of amplitude.

%COV-TF (Equation~\ref{equ:cov-tf}) measures the variability of participants' tapping frequency. IIV (Equation~\ref{equ:iiv}) has also been studied in ~\cite{roalf2018quantitative} showing that AD and MCI groups produced fewer taps with higher IIV compared to health control.

\begin{align}
&\text{M-TF} =\tfrac{1}{K_{p}-1} \sum_{k=2}^{K_{p}} \tfrac{1}{t_{(k)}-t_{(k-1)}} \label{equ:m-tf} \\
&\text{TTC} = min(K_{p}, K_{v}) \label{equ:ttc}\\
&\text{MS} = 1/min(t_{(k)}-t_{(k-1)}) \;  \text{for} \; k=2,3,...,K_{p} \label{equ:ms}\\
&\text{M-ITI} = \tfrac{1}{K_{v}-1} \sum_{k=2}^{K_{v}}(t_{(k)}-t_{(k-1)}) \label{equ:m-iti}\\
&\text{DoS} = \tfrac{1}{K_{p}-1}ln( \tfrac{1}{t_{(2)}-t_{(1)}}/ \tfrac{1}{t_{(K_{p})}-t_{(K_{p}-1)}} ) \label{equ:dos} \\
&\text{COV-A} = \tfrac{\sqrt{(a_{(k)}- \tfrac{1}{K_{p}} \sum_{k=1}^{K_{p}}a_{(k)})^2}}{K_{(p)}}/ (\tfrac{1}{K_{p}} \sum_{k=1}^{K_{p}}a_{(k)}) \label{equ:cov-a}\\
&\text{DoA} = \tfrac{1}{K_{p}}ln \tfrac{a_{(1)}}{a_{(K_{p})}} \label{equ:doa} \\
&\text{COV-TF} = \sqrt{\tfrac{\sum_{k=2}^{K_{p}}( \tfrac{1}{t_{(k)}-t_{(k-1)}} - \text{M-TF})^2}{K_{p}-1}} / \text{M-TF} \label{equ:cov-tf} \\
&\text{IIV} = \sqrt{\tfrac{1}{K_{p}-1} \sum_{k=2}^{K_{p}} [(t_{(k)}-t_{(k-1)})-\text{M-ITI}]^2} \label{equ:iiv}
\end{align}
where $K_{p}$ refers to the number of peaks, $K_{v}$ refers to the number of valleys, $t_{(k)}$ refers to the time point at $k^{th}$ peak and $a_{k}$ refers to the normalized amplitude.

\subsection{Evaluation Method}
We compare the performance of RMT with other state-of-the-art methods, including DLC, to precisely track the thumb-tip and index fingertip during the FTT. The Percentage of Correct Keypoints (PCK) measures the percentage of correctly localized thumb-tip and index fingertip. A correctly localized keypoint is confirmed if the distance (in pixels) between the predicted position and true position is within a pre-set threshold (see equation~\ref{equ:pck}).The Mean of Per Joint Position Error (MPJPE) measures how precisely the predicted position of fingertip can reach compared with the true position (see equation~\ref{equ:mpjpe}). Figure~\ref{fig:pck} shows the PCK of different methods and Table~\ref{tab:mpjpe_comparison} shows the MPJPE of different methods. RMT achieves the best performance in both two metrics.

\begin{align}
&{\text{PCK}}_{@T}=\frac{1}{N\times J}\sum_{j=1}^{J}\sum_{n=1}^{N}(||P^{(j)}_{n}-Y^{(j)}_{n}||_{2}<T) \label{equ:pck}\\
&{\text{MPJPE}}=\frac{1}{N\times J}\sum_{j=1}^{J}\sum_{n=1}^{N}||P^{(j)}_{n}-Y^{(j)}_{n}||_{2} \label{equ:mpjpe}
\end{align}
where $N$ is the number of frames and $J$ is the number of fingertips to be detected (in this case, $J=2$), $P^{(j)}_{n}$ is the predicted position of the $j^{th}$ fingertip on the $n^{th}$ frame, $Y^{(j)}_{n}$ is the true position of the $j^{th}$ fingertip on the $n^{th}$ frame and $T$ is the pre-set threshold.

\section{Results}\label{sec2}
In this section, we show that RMT achieves accurate digit tracking from 2D videos and accurate movement feature extraction.

\subsection{Digit Tracking Results}
\begin{figure}[htbp]
\centering
\includegraphics[width=0.8\textwidth]{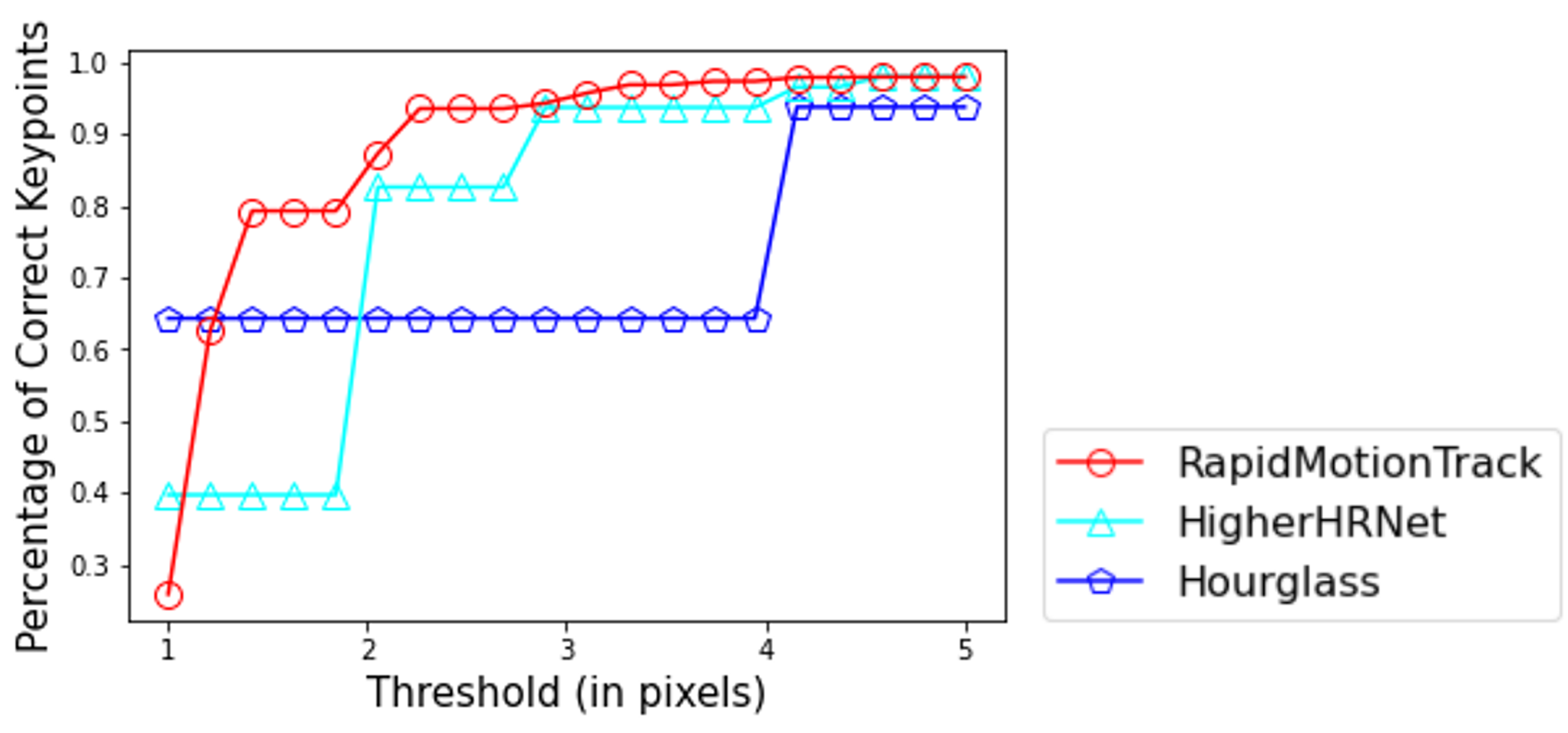}
\caption{Different models' PCK on FTT testing dataset (DLC software does not report PCK). This shows that RMT has the highest percentage of detected correct keypoints at different thresholds when compared to the wearable sensor data (ground truth).}
\label{fig:pck}
\end{figure}

Digit tracking by RMT is more accurate than other available state-of-the-art methods. This is reflected in a higher Percentage of Correct Keypoints (PCK) and a lower Mean of Per Joint Position Error (MPJPE) on the testing dataset shown in Figure~\ref{fig:pck} and Table~\ref{tab:mpjpe_comparison}. The calculations of PCK and MPJPE are described in the Online Method section.

\begin{table}[ht]
\centering
\caption{Different computer vision models' Mean of Per Joint Position Error on FTT testing dataset}
\scriptsize
\begin{tabular}{lccc}
\hline
\textbf{Methods} & \textbf{Thumb-tip} & \textbf{Index Fingertip} & \textbf{Average} \\
\hline
\hline
\textbf{RapidMotionTrack} & \textbf{1.10} & \textbf{1.32} & \textbf{1.21} \\
DLC-ResNet50~\cite{arias2012validity} & -  & - & 3.00 \\
DLC-MobileNet~\cite{arias2012validity} & - & - & 2.30 \\
HigherHRNet~\cite{cheng2020higherhrnet} & 1.47 & 1.71 & 1.59 \\
Hourglass~\cite{newell2016stacked} & 1.37 & 1.82 & 1.60 \\
\hline
\end{tabular}
\label{tab:mpjpe_comparison}
\end{table}

\begin{figure*}[htbp]
%\vskip -.2cm
\centering
\includegraphics[width=1\textwidth]{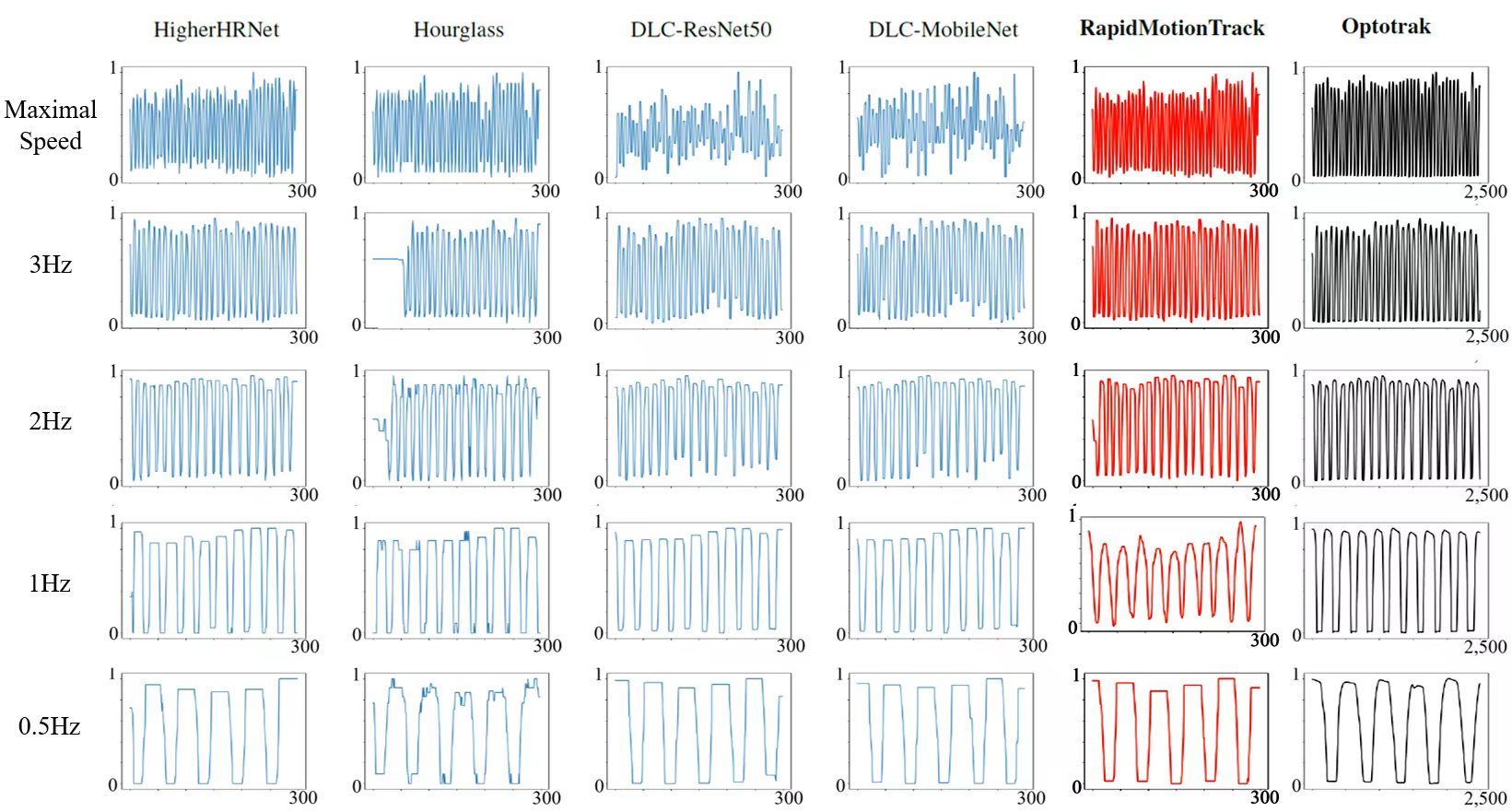}
\caption{Normalized sample distance vs time graphs calculated from computer vision methods in one individual performing the FTT. The blue graphs show the results from each computer vision method compared to the wearable sensor gold standard method, Optotrak (black graph). The X-axis shows the time (frame by frame where 300 frames for computer vision methods and 2,500 frames for the Optotrak method) and the Y-axis shows normalized distance (measured in pixels for computer vision methods and in mm for the Optotrak method). Each row denotes a different FTT condition, where Row 1 is the self paced maximal speed condition and Rows 2, 3, 4 and 5 are the 3Hz, 2Hz, 1Hz and 0.5Hz metronome-paced conditions respectively.}
\label{fig:xyplot}
\end{figure*}

The distance between detected index fingertip and thumb-tip positions normalized to maximum aperture at each frame over time is plotted in Figure 2 (Figure~\ref{fig:xyplot}). A visual comparison to the wearable sensor method demonstrates that when applied to the unlabelled videos, RMT provides good tracking performance in the `maximal speed' condition (approximately 6Hz).

\newpage

\subsection{Feature Extraction Results}
The RMT system, HigherHRNet and Hourglass models can extract valid hand movement features even when participants tap in a high frequency range ($>$4Hz). In comparison, hand movement speed, amplitude, and rhythm related features calculated based on the DLC system's digit tracking result are invalid when participants tap more quickly than 4Hz. These are reflected in the comparison between different computer vision methods and the Optotrak in terms of Bland Altman plots, X-Y plots (Figure~\ref{fig:ba-corr}) and statistical tests (Table~\ref{tab:validation_study_welch_ttest}). Details of movement features are explained on the Online Method section.

\iffalse
\begin{figure*}[htbp]
%\vskip -.2cm
\centering
\begin{tabular}{cccc}
\scriptsize \textbf{M-TF} & \scriptsize \textbf{Logarithm of COV-TF} & \scriptsize \textbf{Logarithm of COV-A}
\\
\includegraphics[scale=0.15]{fig_ba_M-TF.png} &
\includegraphics[scale=0.15]{fig_ba_COV-TF.png}&
\includegraphics[scale=0.15]{fig_ba_COV-A.png}
\\ 
\scriptsize \textbf{TTC} & \scriptsize \textbf{M-ITI} & \scriptsize \textbf{Logarithm of IIV} \\
\includegraphics[scale=0.15]{fig_ba_TTC.png}&
\includegraphics[scale=0.15]{fig_ba_M-ITI.png}&
\includegraphics[scale=0.15]{fig_ba_IIV.png}
\\
\scriptsize \textbf{MS} & \scriptsize \textbf{Logarithm of DoA} & \scriptsize \textbf{Logarithm of DoS} \\
\includegraphics[scale=0.15]{fig_ba_MS.png}&
\includegraphics[scale=0.15]{fig_ba_DoA.png}&
\includegraphics[scale=0.15]{fig_ba_DoS.png}
\\
&&\includegraphics[scale=1]{fig_legend.png}
\\
\end{tabular}
\vskip -.2cm
\caption{Bland Altman Plots of different features calculated by different computer vision methods (Y-axis) vs the Optotrak method (X-axis). M-TF = mean tapping frequency, COV-TF = coefficient of variance of tapping frequency, COV-A = coefficient of variance of amplitude, TTC = total tap count, M-ITI = mean inter tap interval, IIV = intra-individual variability, MS = maximum speed, DoA = decrement of amplitude, DoS = decrement of speed.}
\label{fig:ba}
\end{figure*}
\fi

\begin{figure*}[htbp]
\centering
\includegraphics[scale=0.4]{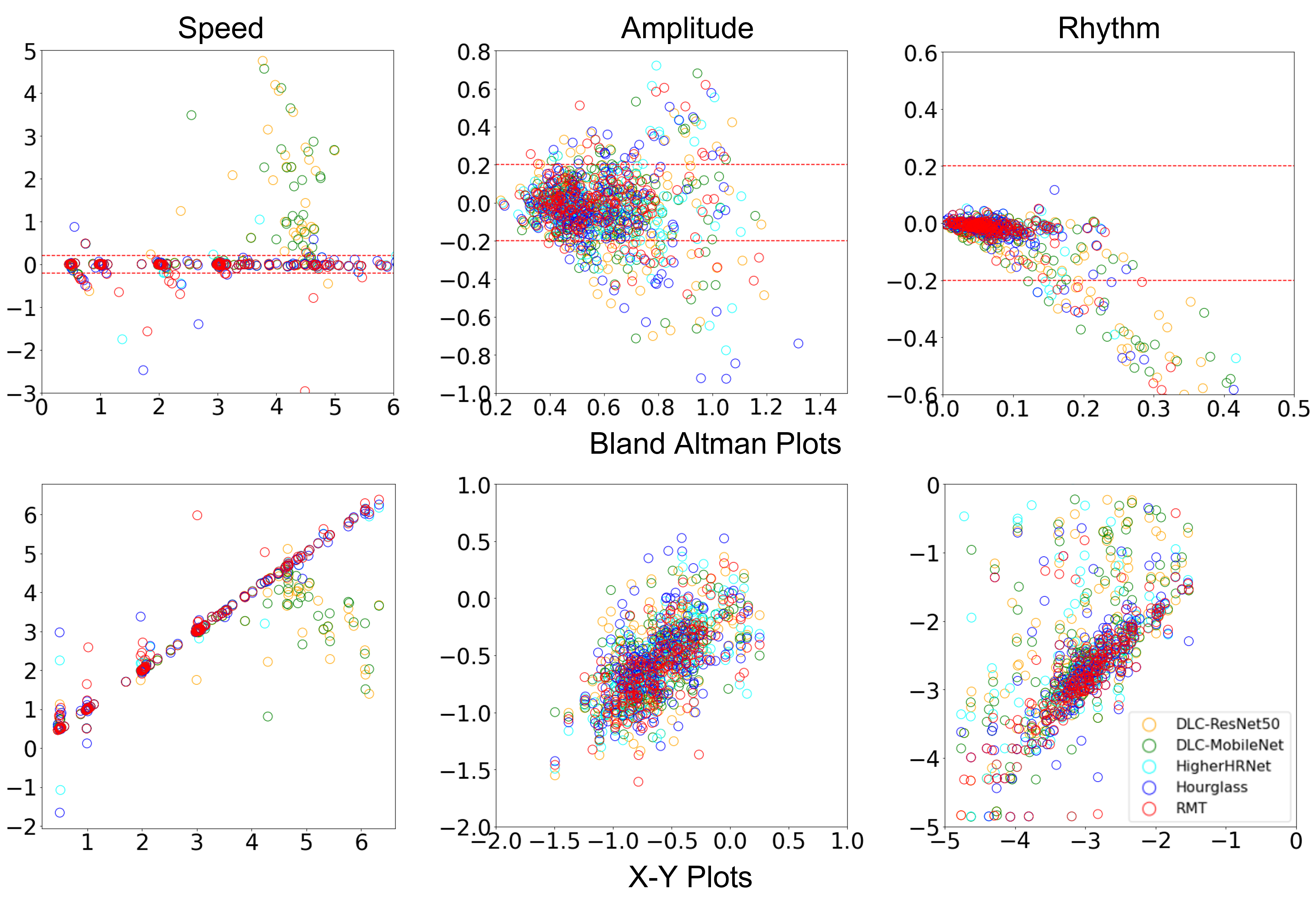}
\caption{Features comparison: computer vision method (Y-axis) vs Optotrak method (X-axis). The top row presents Bland Altman Plots of different features calculated by computer vision methods (Y-axis) vs the Optotrak method (X-axis). The bottom row presents X-Y Plots of different features calculated by computer vision method (Y-axis) vs the Optotrak method (X-axis). Speed is calculated from the mean of tapping frequency, amplitude is calculated from by the logarithm of the coefficient of variance of normalized tapping amplitude and rhythm is calculated by the logarithm of the coefficient of variance of tapping frequency. Detailed comparisons of other features are shown in the supplementary material.}
\label{fig:ba-corr}
\end{figure*}

\begin{table}[htbp]
\centering
\caption{Welch's t test (at 0.05 significant level) results of different computer vision methods compared with the Optotrak measures on speed, amplitude and rhythm. Results from two methods have no significant difference when $P>0.05$ (accept null hypothesis), and have significant difference when $P<0.05$ (reject null hypothesis).}

\begin{tabular}{lllll}
\hline
\textbf{Condition} & \textbf{Computer Vision Method} & \textbf{Speed} & \textbf{Amplitude} & \textbf{Rhythm} \\
\hline
& RMT & accept & \textbf{accept} & accept  \\
Maximal speed ($<$4Hz) & DLC-ResNet50 & accept & reject & accept \\
& DLC-MobileNet & accept & reject & accept  \\
\hline
& RMT & \textbf{accept} & \textbf{accept} & accept \\
Maximal speed ($>$4Hz) & DLC-ResNet50 & reject & reject & accept \\
& DLC-MobileNet & reject & reject & accept \\
\hline
& RMT & accept & accept & accept \\
3Hz & DLC-ResNet50 & accept & accept & accept \\
& DLC-MobileNet & accept & accept & accept \\
\hline
& RMT & accept & accept & accept \\
2Hz & DLC-ResNet50 & accept & accept & accept \\
& DLC-MobileNet & accept & accept & accept \\
\hline
& RMT & accept & accept & accept \\
1Hz & DLC-ResNet50 & accept & accept & accept \\
& DLC-MobileNet & accept & accept & accept  \\
\hline
& RMT & accept & accept & accept \\
0.5Hz & DLC-ResNet50 & accept & accept & accept \\
& DLC-MobileNet & accept & accept & accept  \\
\hline
\end{tabular}
\label{tab:validation_study_welch_ttest}
\end{table}

\iffalse
\begin{table}[htbp]
\centering
\caption{Welch's t test (at 0.05 significant level) results of different computer vision methods compared with the Optotrak measures on speed, rhythm and decrement. The two methods have no significant difference when $P>0.05$ (accept null hypothesis), and have significant difference when $P<0.05$ (reject null hypothesis).}
\begin{tabular}{lllll}
\hline
\textbf{Condition} & \textbf{Methods} & \textbf{Speed} & \textbf{Rhythm} & \textbf{Decrement} \\
\hline
& RMT & \textbf{0.87} & \textbf{0.53} & 0.60 \\
Maximal speed & DLC-ResNet50 & 0 & 0 & 0.35 \\
& DLC-MobileNet & 0 & 0 & 0.26 \\
\hline
& RMT & 0.83 & \textbf{0.7} & 0.70  \\
Maximal speed ($<$4Hz) & DLC-ResNet50 & 0.94 & 0 & 0.48 \\
& DLC-MobileNet & 0.98 & 0 & 0.99  \\
\hline
& RMT & \textbf{0.88} & \textbf{0.59} & 0.96 \\
Maximal speed ($>$4Hz) & DLC-ResNet50 & 0 & 0 & 0.43 \\
& DLC-MobileNet & 0 & 0 & 0.92 \\
\hline
& RMT & 0.27 & 0.59 & 0.69 \\
0.5Hz & DLC-ResNet50 & 0.53 & 0.41 & 0.33 \\
& DLC-MobileNet & 0.88 & 0.37 & 0.49  \\
\hline
& RMT & 0.98 & 0.96 & 0.44 \\
1Hz & DLC-ResNet50 & 0.87 & 0.99 & 0.80 \\
& DLC-MobileNet & 0.82 & 0.44 & 0.60  \\
\hline
& RMT & 0.20 & 0.5 & 0.81 \\
2Hz & DLC-ResNet50 & 0.81 & 0.65 & 0.67 \\
& DLC-MobileNet & 0.94 & 0.75 & 0.30 \\
\hline
& RMT & 0.35 & 0.56 & 0.83 \\
3Hz & DLC-ResNet50 & 0.16 & 0.69 & 0.55 \\
& DLC-MobileNet & 0.20 & 0.59 & 0.27 \\
\hline
\end{tabular}
\label{tab:validation_study_welch_ttest}
\end{table}
\fi

\newpage

\subsection{Generalisability of RMT to other fast human movements}
RMT was also able to accurately track a range of other human movements when applied to webcam videos of head turning, toe tapping~\cite{enoki2019foot}, leg agility, and sit-to-stand movements~\cite{whitney2005clinical,duncan2011five} as shown in Figure~\ref{fig:other}. The tracking videos are provided in the supplementary materials.

\begin{figure*}[htbp]
\centering
\includegraphics[width=\textwidth]{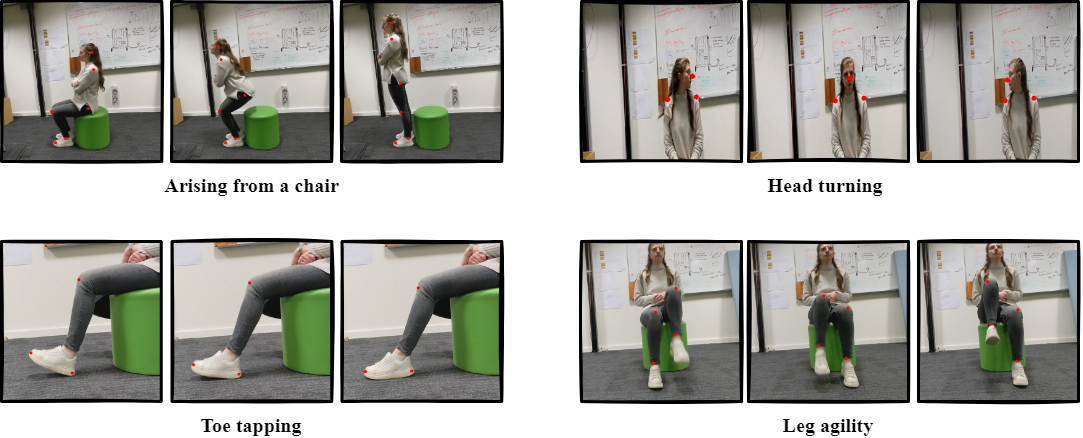}
\caption{RMT accurately tracks a range of rapid human motions including: head turning, arising from a chair, toe tapping and leg agility. The keypoints are denoted by red dots and the video in supplementary materials demonstrates these keypoints track throughout the motion.}
\label{fig:other}
\end{figure*}    

\section{Discussion and Conclusion}

\subsection{Precise Digit Tracking from 2D Laptop Camera Videos}
Digit tracking by RMT is more accurate than available state-of-the-art methods. When participants tap in a fast range (e.g. higher than 4Hz), DLC cannot accurately and reliably track the digits~\cite{li2022moving}. Additionally, this will further lead to incorrect digit distance vs time graph. Figure~\ref{fig:xyplot} demonstrated an individual participant’s distance (between index fingertip and thumb-tip) vs time graphs under 5 conditions of finger tapping calculated from different computer vision methods and the Optotrak method. In the maximal speed condition, motion blur frames were very common. Compared with other deep learning networks to detect keypoints, the distance vs time graph obtained from the RMT system is more stable and closer to the gold standard Optotrak methods. This is due to the P-MSDSNet fusing multi-scale features from both same depth levels and different depth levels in a parallel manner, which can gradually refine the features and detected fingertip positions layer by layer and scale by scale even on a blurred frame. An accurate digit tracking lays the foundation for extracting hand movement features. 

\subsection{Feature Extraction Discussion}
Despite the accurate digit tracking results, the unsmooth distance vs time curve may also affect the accuracy of final feature extraction. To overcome this issue, the proposed Adaptive Vertex Recognition algorithm can help RMT system correctly identify the peaks and troughs from a particular distance vs time graph without being affected by the noise or unsmooth part.

\subsection{Strengths, Limitations and Future Direction}
A particular strength of RMT is that it takes original finger tapping videos as input, through the Fingertip Tracking Module and Adaptive Vertex Recognition Module, and outputs accurate and valid hand movement features for both assessment and research. This is an additional output compared with DLC which only tracks digits (RMT not only tracks digits but also produces valid movement feature data). In addition, both digit tracking and the feature extraction achieve the best results out of all currently available state of the art computer vision deep learning models and this can serve for not only FTT but also other movement test analysis including sit-to-stand movement test~\cite{whitney2005clinical,duncan2011five} and foot-tapping test~\cite{enoki2019foot} from recorded 2D videos. We showed a comparison of RMT with the DLC and the Optotrak system in Table~\ref{tab:compare}; this shows that the features extracted using RMT are equivalent to wearable sensor methods but, in addition, have the potential to be performed remotely (e.g., for telemedicine) and at the population level (due to the wide reach of webcams). Rather than the Optotrak system and DLC whose output is fingertip position, the RMT system can directly output validated finger tapping features. 

\begin{table}[ht]
\centering
\caption{Output data from RMT , DLC and wearable sensor method}
\scriptsize
\begin{tabular}{lccc}
\hline
 & \textbf{RMT system} & \textbf{DLC} &  \textbf{Wearable sensor methods} \\
\hline
\hline
%Remoteness & Yes & Yes & No \\
Finger Digits Tracking & Yes & Yes & Yes \\
Feature Extraction & Yes & No & Yes \\
Feature Validity & Yes & No & Yes \\
\hline
\end{tabular}
\label{tab:compare}
\end{table}

Currently RMT is offline but future work will develop this into an online system. Additionally, we will extend RMT that can be used for general human movement feature extraction. First, we will build a model zoo consisting of different models for different movements, and keep updating the zoo. Second, we will make RMT as a windows-based system that can be used in a more efficient manner for neuroscience. RMT will also be used in the TAS Test project~\cite{alty2022tas}, which aims to detect the earliest stages of Alzheimer's disease using hand movement analysis.

\bibliographystyle{unsrt}
\bibliography{sample}

\end{document}